\documentclass[conference]{IEEEtran}
\IEEEoverridecommandlockouts
\usepackage{cite}
\usepackage{amsmath,amssymb,amsfonts}
\usepackage{algorithmic}
\usepackage{graphicx}
\usepackage{textcomp}
\usepackage[usenames, dvipsnames]{xcolor}
\usepackage{makecell}
\usepackage{multirow}
\usepackage{microtype}
\usepackage{array}
\usepackage{booktabs}
\usepackage{times}
\usepackage{latexsym}
\usepackage{stfloats}
\usepackage{marvosym}
\usepackage{tabularx}

\def\BibTeX{{\rm B\kern-.05em{\sc i\kern-.025em b}\kern-.08em
    T\kern-.1667em\lower.7ex\hbox{E}\kern-.125emX}}
\begin{document}

\title{\huge{KEPR: Knowledge Enhancement and Plausibility Ranking for Generative Commonsense Question Answering}
\thanks{\textsuperscript{\Letter}Corresponding author.}
\thanks{The research is supported by the National Key R\&D Program of China (2020YFB1313601) and the National Science Foundation of China (No.62076174, No.61836007).}}


\author{
\IEEEauthorblockN{Zhifeng Li}
\IEEEauthorblockA{
\textit{School of Computer Science and Technology} \\
\textit{Soochow University}, Suzhou, China \\
li\_2hifeng@outlook.com}
\\
\IEEEauthorblockN{Yifan Fan}
\IEEEauthorblockA{
\textit{School of Computer Science and Technology} \\
\textit{Soochow University}, Suzhou, China \\
yifanfannlp@gmail.com}
\and 
\IEEEauthorblockN{Bowei Zou}
\IEEEauthorblockA{
\textit{Institute for Infocomm Research} \\
\textit{A*STAR}, Singapore \\
zou\_bowei@i2r.a-star.edu.sg}
\\
\IEEEauthorblockN{Yu Hong\textsuperscript{\Letter}}
\IEEEauthorblockA{
\textit{School of Computer Science and Technology} \\
\textit{Soochow University}, Suzhou, China \\
tianxianer@gmail.com
}}

\maketitle

\begin{abstract}
Generative commonsense question answering (GenCQA) is a task of automatically generating a list of answers given a question. The answer list is required to cover all reasonable answers. This presents the considerable challenges of producing diverse answers and ranking them properly. Incorporating a variety of closely-related background knowledge into the encoding of questions enables the generation of different answers. Meanwhile, learning to distinguish positive answers from negative ones potentially enhances the probabilistic estimation of plausibility, and accordingly, the plausibility-based ranking. 
Therefore, we propose a Knowledge Enhancement and Plausibility Ranking (KEPR) approach grounded on the Generate-Then-Rank pipeline architecture. Specifically, we expand questions in terms of Wiktionary commonsense knowledge of keywords, and reformulate them with normalized patterns. Dense passage retrieval is utilized for capturing relevant knowledge, and different PLM-based (BART, GPT2 and T5) networks are used for generating answers. On the other hand, we develop an ELECTRA-based answer ranking model, where logistic regression is conducted during training, with the aim of approximating different levels of plausibility in a polar classification scenario. 
Extensive experiments on the benchmark ProtoQA show that KEPR obtains substantial improvements, compared to the strong baselines. Within the experimental models, the T5-based GenCQA with KEPR obtains the best performance, which is up to 60.91\% at the primary canonical metric Inc@3. It outperforms the existing GenCQA models on the current leaderboard of ProtoQA.
\end{abstract}


\section{Introduction}

Great efforts were made on the study of Commonsense Question Answering (CQA) in the area of natural language processing. The conventional CQA tasks can be roughly divided into two categories, including assertion correctness judgment~\cite{talmor2022commonsenseqa} and multiple-choice question answering~\cite{talmor2019commonsenseqa, ghosal-etal-2022-cicero}. Both stimulate the exploration of discriminative CQA methods, as well as supportive neural models. Different from the aforementioned CQA tasks, Generative CQA (GenCQA) produces answers in a generative way, as claimed in the task definition towards the benchmark ProtoQA~\cite{boratko2020protoqa}. In particular, models necessarily generate diverse answers via commonsense reasoning within a specific prototypical scenario and rank them in terms of plausibility (see the example in Table~\ref{tab1:exampleprotoqa}).

\begin{table}[t]
\begin{center}
\caption{A GenCQA case which is selected from ProtoQA.} 
\renewcommand{\arraystretch}{1.0}

\begin{tabular}{cc}
\hline
\specialrule{0em}{0.5pt}{0.5pt}
\makecell[c]{Prototypical\\ Question} &
\makecell[l]{Name something that an \textit{\textbf{athlete}} would not\\
keep in her \textit{\textbf{refrigerator}}.}\\
\specialrule{0em}{0.5pt}{0.5pt}
\hline
\specialrule{0em}{0.5pt}{0.5pt}
\makecell[c]{Ground-truth\\answer list} & \makecell[l]{
unhealthy food (36):  chocolate, junk food, ...\\
unhealthy drinks (24):  coke, alcohol, ...\\
clothing/shoes (24):  gloves, clothes, shoe, ...\\
accessories (7):  handbag, medal, tennis, ...\\
}\\
\specialrule{0em}{0.5pt}{0.5pt}
\hline
\specialrule{0em}{0.5pt}{0.5pt}
[BART] & \makecell[l]{food, ice, vitamins, beer, alcohol}\\
\specialrule{0em}{0.5pt}{0.5pt}
\hline
\specialrule{0em}{0.5pt}{0.5pt}
\makecell[c]{Background\\ Knowledge} & 
\makecell[l]{
\textit{\textbf{athlete}}: A person who actively participates in \\ physical sports, especially with great skill.\\
\textit{\textbf{refrigerator}}: A household appliance used for \\ keeping food fresh by refrigeration.}\\
\specialrule{0em}{0.5pt}{0.5pt}
\hline
\specialrule{0em}{0.5pt}{0.5pt}
\multicolumn{2}{l}{\makecell[l]{
\specialrule{0em}{0.5pt}{0.5pt}
The ground-truth answers are organized into different classes,\\
conditioned on plausibility scores. [BART] corresponds to the\\
generated results by BART-based, GenCQA without consider-\\
ing the knowledge of ``{\em athlete}'' and ``{\em refrigerator}''.\\
}}
\end{tabular}
\label{tab1:exampleprotoqa}
\end{center}
\end{table}

Commonsense-aware neural models contribute to the solution of CQA problems. The pre-trained generative language models such as GPT2~\cite{radford2019language}, BART~\cite{lewis2020bart} and T5~\cite{raffel2020exploring} are knowledgeable, and therefore serve as highly competitive baselines for CQA. However, empirical findings show that they fail to perform perfectly in the GenCQA task, where the answers they produced are of less diversity and generally disordered. For example, within the five answers generated by a fine-tuned BART (in Table~\ref{tab1:exampleprotoqa}), ``{\em beer}'' appears as a redundant case when ``{\em alcohol}'' has been generated, and there are merely one class of plausible answers produced, i.e., ``{unhealthy drinks}''. More seriously, the incorrect answers such as ``{\em ice}'' and ``{\em food}'' are ranked higher than plausible answers.

We suggest that the above issues can be alleviated by the following two solutions. First, the supplementation of exclusive background knowledge of keywords in the questions is beneficial for pursuing a wider range of answers. For example, the nature of ``{\em athlete}'' like ``{\em participating in physical sports}'' (see the row of Background Knowledge in Table~\ref{tab1:exampleprotoqa}) is informative for sampling the answer ``{\em junk food}'', during the evidential reasoning process. Similarly, the attributes of ``{\em refrigerator}'' are useful for inferring the answer classes ``{\em clothing/shoes}'' and ``{\em accessories}''. Second, it is crucial to construct a separate ranking mechanism conditioned on the probabilistic estimation of plausibility, instead of the direct use of beam search at the decoding stage.

Accordingly, we propose a Knowledge Enhancement and Plausibility Ranking (abbr., KEPR) approach\footnote{Access code and appendix via https://github.com/Zaaachary/CSQA-KEPR} for the GenCQA task. KEPR is grounded on the generate-then-rank~\cite{shen2022joint} framework, which decouples the answer generation and ranking tasks. For answer generation, we enhance the PLM-based generative model by incorporating commonsense knowledge of keywords into the modeling process, where Wiktionary\footnote{https://www.wiktionary.org/} is used. A series of easy-to-implement assistive technologies are applied, including statistical keyword extraction, dense passage retrieval and pattern-based question rewriting. For answer ranking, we construct an ELECTRA-based~\cite{clark2020electra} ranker which estimates the plausibility of answers with logistic regression. The ranker is separately trained in a binary classification scenario using the most plausible answers and randomly-selected negative cases. Besides, answer deduplication is used.

We experiment on the benchmark corpus ProtoQA, and the experimental results demonstrate that KERP is effective and multi-model compatible. The main contributions of this paper are as follows:
\begin{itemize}
    \item We propose a novel method, i.e., KEPR, to enhance GenCQA, where knowledge enhancement and plausibility ranking are used. KEPR produces considerable improvements at all canonical metrics, compared to GPT2, BART and T5, as well as their robust versions.
    \item When cooperating KEPR with BART, a relatively-weak PLM, we achieve a comparable performance to the state-of-the-art GenCQA model~\cite{maskedLuo22} for Inc@3 (55.38\% vs 55.77\%) but with a smaller model size.
    \item When cooperating KEPR with a strong PLM like T5-3B, we achieve the best performance, where the Inc@3 score is up to 60.91\%.
\end{itemize}


\section{Related Works}

\textbf{Generative CQA}\quad Ma et al.~\cite{ma2021exploring} use fine-tuning, auto-prompting and prefix-tuning to enhance PLM-based GenCQA models. Their systematical analysis reveals that fine-tuning leads to the forgetting of global knowledge gained during pre-training. This is unavoidable due to the representational approximation or even overfitting to the task-specific data. Luo et al.~\cite{maskedLuo22} develop a popularity-aware answer ranker. It is not only used for re-ranking the generated answers in practice, but serves as a guider to refine the generator for producing typical answers. Policy gradient based rewarding is utilized to reinforce the training of answer generator.

\textbf{Commonsense Generation}\quad The studies of concept-oriented commonsense description generation on CommonGen~\cite{lin2020commongen} have been widely conducted. They can be safely considered as references because the ProtoQA~\cite{boratko2020protoqa} and CommonGen projects share the same goal, i.e., verifying linguistic intelligence from the perspective of generative commonsense reasoning. In this field, research effort has been devoted to the incorporation of external knowledge into generative language models. Specifically, EKI-BART~\cite{fan2020enhanced} retrieves prototypes from in-domain and out-of-domain corpora, and utilizes them as the scenario knowledge to guide the generation process. KFCNet~\cite{li-etal-2021-kfcnet-knowledge} acquires high-quality prototypes and applies contrastive learning to constrain both intrinsic representations of encoder and extrinsic representations of decoder. KG-BART~\cite{liu2021kg} utilizes a knowledge graph to augment BART for text generation, where concept relationships are incorporated. Moreover, it applies graph attention for semantic enrichment.

\textbf{Discriminative CQA}\quad It is noteworthy that a variety of discriminative CQA models~\cite{banerjee2019careful, talmor2019commonsenseqa,talmor2022commonsenseqa} leverage external knowledge to implement the deeper reasoning. A wider range of knowledge bases is considered, such as ConceptNet~\cite{lin2019kagnet}, Cambridge Dictionary~\cite{chen2020improving}, Open Mind Common Sense~\cite{li2021winnowing} and Wiktionary~\cite{xu2021human}. Unfortunately, such sophisticated techniques cannot be transferred to the generative CQA task. It is because that they heavily rely on the exposed answer candidates for retrieving and modeling relevant external knowledge. Instead, all possible answers in GenCQA aren't given, but on the contrary, they need to be generated eventually. 

\begin{figure*}[t]
\centering
\includegraphics[width = 0.9\linewidth]{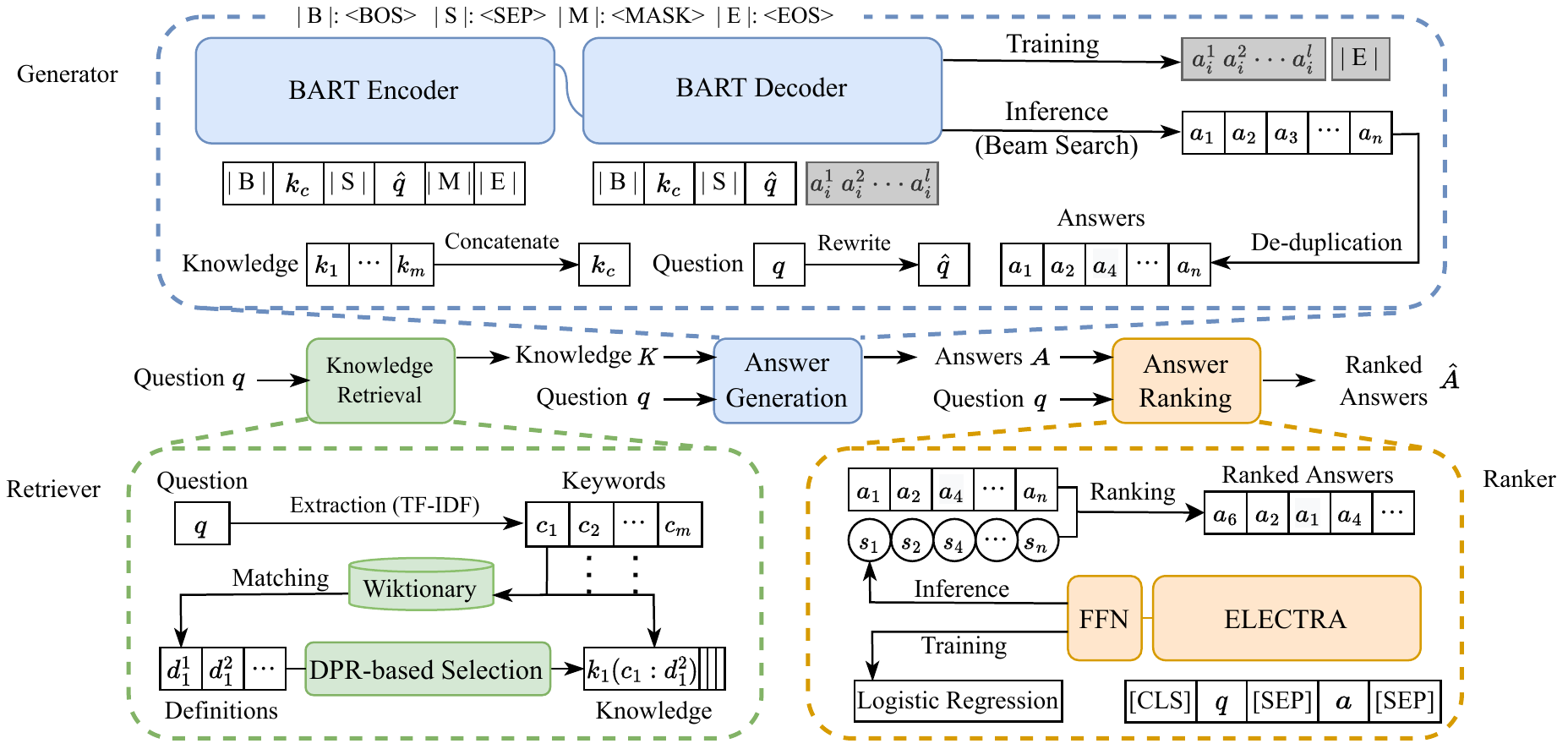}
\caption{The overall architecture of KEPR. The grey blocks in Answer Generation show the input and output of each generation process, where $a_i^l$ denotes the $l$-th token of the $i$-th answer of the given question.
}
\label{fig1:overall_approach}
\end{figure*}

\section{APPROACH}

The overall architecture of KEPR-based GenCQA model is shown in Fig.~\ref{fig1:overall_approach}. The considered PLM in this case is BART. It can be replaced by other PLMs, such as GPT2 and T5. The GenCQA model consists of three components, including knowledge-oriented retriever, answer generator and ranker.

\textbf{Retriever} first extracts keywords from the question $q$ using statistical information. Dense passage retrieval is further used to acquire relevant knowledge $K=\{k_1, k_2, ...,k_m\}$ of the keywords from Wiktionary, where $k_i$ denotes the knowledge item of the $i$-th keyword. Each knowledge item describes the nature of a keyword in a sentence.

\textbf{Generator} learns to generate different answers $A = \{a_1, a_2, ..., a_n\}$ conditioned on the question $q$ and relevant knowledge items. The BART-based encoder is used for knowledge-aware semantic feature representation. The accompanying decoder infers the possible answers in terms of the features. Question rewriting and answer deduplication are used to assist the generator.

\textbf{Ranker} learns to rate the generated answers with the real-valued plausibility scores. Each score is normalized and signals the probability that a certain answer is commonly recognized to be reasonable. The ranker is constructed with ELECTRA. It is separately trained to compute plausibility scores, where logistic regression is conducted within the binary classification scenario of truly-plausible instances and negative ones.

\subsection{Knowledge Acquisition (Retriever)}
We retrieve knowledge items using the keywords of the question $q$ as queries. The large-scale online dictionary Wiktionary is used as the knowledge source, which contains about 7M English sentence-level knowledge items. We extract the top-$m$ keywords which, statistically, are of higher Term Frequency-Inverse Document Frequency (TF-IDF)~\cite{salton1988term}. For each token $t_j$ in the question $q$, we estimate TF in terms of the occurrence frequency of $t_j$ in $q$, while IDF is calculated using the non-repetitive occurrence frequency of $t_i$ in all the questions in the training set. We lemmatize the keywords by NLTK, and conduct string matching between each keyword and Wiktionary entries to acquire the corresponding knowledge item. 

However, multiple knowledge items can be obtained for each keyword. It is because that, in Wiktionary, one entry (e.g., a polysemous word) generally corresponds to multiple knowledge items, which specify different definitions and natures. In order to obtain the knowledge item that solely applies to the question $q$, we use the Dense Passage Retriever (DPR) ~\cite{karpukhin2020dense} to perform context-aware text matching. Specifically, we reformulate the query by the fixed prototype as below: ``\emph{What is the meaning of word \textless keyword\textgreater in the sentence \textless question\textgreater }?''. The slots ``\textless keyword\textgreater'' and ``\textless question\textgreater'' can be dynamically filled with concrete instances. On the basis, we use DPR to respectively encode the query and each candidate knowledge item. This allows their sentence-level embeddings to be produced. DPR further calculates the relevant score over embeddings using the scalar product. The knowledge item that holds the highest relevance score will be adopted, while the rest will be abandoned. 

Accordingly, we construct an exclusive knowledge set $K=\{k_1, k_2, ..., k_m\}$ for each question $q$, where $k_i$ is the concatenation of $i$-th keyword and its relevant knowledge item. 

\subsection{Knowledge-aware Answer Generation (Generator)}

\subsubsection{Question Rewriting}

We rewrite the questions to produce {Cloze}-style equivalents that contain the masked suffixes. We provide an example below, where the sequence in \textcolor{Salmon}{orange} font serves as the prefix which is rewritten, while the sequence in \textcolor{Green}{green} font appears as the main body text which isn't falsified.
\begin{itemize}
    \item [] {Original Question: {\em \textcolor{Salmon}{Name something} \textcolor{Green}{that an athlete would not keep in her refrigerator}}}?
    
    {Rewritten Question: {\em \textcolor{Salmon}{One thing} \textcolor{Green}{that an athlete would not keep in her refrigerator }} is \underline{\textless MASK\textgreater}?}
\end{itemize}
The rewritten questions, ideally, are PLM-friendly because text infilling (masked span prediction) is generally considered as the task for pre-training. A series of predefined patterns are utilized for rewriting questions. They can be easily ascertained and invoked by looking up the prefix-pattern mapping table, as shown in Appendix A.

\subsubsection{Knowledge-enhanced Generation} 
To demonstrate the input and output of our knowledge-enhanced answer generator, we chose BART as an example here, which is based on the encoder-decoder architecture. The input of BART encoder is constructed by concatenating knowledge items with the rewritten question, in the format as below: 
\begin{equation}
\small 
\mathrm{<BOS> k_c <SEP> \hat{q} <MASK> <EOS>}
\end{equation}
where, $k_c$ denotes the concatenated sentence-level knowledge items, i.e., the ones in $K$, while $\hat{q}$ denotes the rewritten question. The input of BART decoder is constructed in the similar way except that the special tokens {\small \textless MASK\textgreater} and {\small \textless EOS\textgreater} are pruned, since they are the generation target of decoder.

We also consider GPT2 and T5 for answer generation in our experiments. GPT2 is a decoder-only model that performs unidirectional autoregressive generation. When using GPT2, we adopt the decoder methodology used by BART, which generates answers by predicting the next token based on the knowledge and question. T5 is structurally similar to BART, and we use it in a way that is similar to BART, except that T5 uses a special token, \textless extra\_id \textgreater, to mask the answer.

\subsubsection{Training Stage} We train the generator in the teacher-forcing manner~\cite{arora2022exposure}, where the truly-correct preceding context (i.e., $t$-1 ground-truth tokens) are necessarily exposed to the generator when the $t$-th token is being predicted. The training objective is to maximize the global transition probability $P_{\theta}$ as below: 
\begin{equation}
\small
\begin{array}{l}
\begin{aligned}
&\boldsymbol\theta=argmax. \sum_{q_i\in Q}\sum_{a_j\in A_i} P_{\theta}(a_j|\hat{q}_i,\boldsymbol{K}_i)\\
&P_{\theta}(a_j|\hat{q}_i,\boldsymbol{K}_i)=\sum_{t=1}^{len}P_{\theta}(a_j^t|\hat{q}_i,\boldsymbol{K}_i)
\end{aligned}
\end{array}
\label{formula1:process}
\end{equation}
where $\theta$ denotes all the parameters of the generator. $K_i$ is the knowledge set of the $i$-th question $q_i$, while $a_j^t$ is the $t$-th token of the $j$-th answer of $q_i$.

\subsubsection{Inference Stage} We sample a fixed number of answer candidates for each question during inference stage, where beam search~\cite{cho2014properties} is used. By beam search, the answers which are of higher confidence will be adopted. The sum of logarithmic generation probabilities of tokens is used for estimating the confidence. Note that there might be certain different forms of the same answer in the list. To address the duplication issue and obtain diverse answers that possess different senses, we conduct dictionary-based deduplication. Firstly, we remove stop words from the answer candidates and perform lemmatization, so as to convert them into content words which are of original morphology. For example, given the list of answer candidates like ``{\em a bike}'', ``{\em her bikes}'' and ``{\em the bicycle}'', we convert them into ``{\em bike}'' and ``{\em bicycle}''.  On this basis, we identify the answer candidates which are thoroughly consisted of synonyms. The NLTK is used for determining synonymy conditioned on WordNet~\cite{miller1995wordnet}. Given a group of synonymous candidates, we merely retain the one holding the highest confidence score and abandon the rest. 

\subsection{Plausibility-based Ranking (Ranker)}
The canonical GenCQA evaluation system additionally evaluates the rankings of the generated answers. Accordingly, if the manually-designated high-ranking answer classes are actually ranked lower, the GenCQA performance will be considered to be less promising. Therefore, we build a ranker to arrange the generated answers in the order of plausibility. We approximate the plausibility by estimating the probability that an answer derives from the class of absolutely positive instances. We separately train a binary classification model (\textbf{P}ositive versus \textbf{N}egative), and use its discriminative layer along with the Sigmoid activation function to compute the probabilistic plausibility score. 

\subsubsection{Training Data Collection}

In order to train the classifier, we construct a training set containing positive and negative instances. All the instances are collected from the ProtoQA training set itself, without using external data. For each question in the ProtoQA training set, we select top-$n$ highly-weighted answers as the absolutely positive instances. The weight is given as the ground truth in ProtoQA, and it is equivalent to the proportion of ``{\em yes}'' vote towards the plausibility of a certain answer category. 

Given a question, we collect negative instances from the answers of other questions, where random sampling is used. To ensure the absolute implausibility, we verify whether the sampled instances are synonymous or contain synonyms with the ground-truth answers. Such cases will be abandoned. In our experiments, we set $n$ to 2, and thus obtain about 16.5K QA pairs of 8,782 positive cases and the same number of negative cases. They are used for training. In the same way, we collect 3.5K instances to build the validation set.

\subsubsection{Plausibility Approximation}
The binary classification model is constructed with the ELECTRA-based encoder~\cite{clark2020electra} and a fully-connected linear layer with Sigmoid. We simply use the latter as the discriminator. For a question $q$ and one of the generated answers $a$, we concatenate them to form the input of ELECTRA encoder: ``[CLS] $q$ [SEP] $a$ [SEP]''. Given the output of ELECTRA, we take the encoded global representation [CLS], and feed it into the discriminator. On this basis, we use the linear layer to project [CLS] into the $1$-dimensional vector $v\in \mathbb{R}$ that represents the plausibility level. Sigmoid function is further used to activate $v$, producing a probabilistic value $\hat{v}$. We sort all the generated answers of $q$ in terms of $\hat{v}$. The whole classifier is trained using the aforementioned training set (Section III.C.1), in the manner of logistic regression. Minimizing the binary cross-entropy loss is considered as the objective: 
\begin{equation}
\small
	\begin{split}
	\mathcal{L}(q,a) = - \sum\ y \cdot log(\hat{v}) + (1-y) \cdot log{(1-\hat{v})}
	\end{split}
\end{equation}

Through the knowledge-enhanced answer generation, we deliberately retain 12 answers in total for each question. The number (12) is larger than the specified maximum amount (10) of answers in the task of GenCQA. In practice, we utilize the Ranker to evaluate the plausibility of every generated answer and arrange them based on their plausibility, subsequently eliminating the two answers that rank the lowest.

\section{Experimentation}

\subsection{Corpus and Datasets}

We carry out experiments on ProtoQA~\cite{boratko2020protoqa}, a benchmark corpus of GenCQA. ProtoQA comprises a training set and a test set as usual, though it provides two validation sets, namely VSet1 and VSet2 for short. VSet1 (also known as scraped Dev) is built in the same way as the training set, where the answers of each question are (1) collected by questionnaire survey, (2) weighted by voting scores and (3) purified by automatic deduplication, without being double-checked and classified. By contrast, the questions in both VSet2 (also known as crowd-sourced Dev) and test set are answered by annotators from multiple perspectives, and the answers are carefully verified, ranked and classified. The statistics are shown in Table~\ref{tab2:dataset}.

\begin{table}[th]
\begin{center}
\caption{The statistics of ProtoQA datasets.}
\label{tab2:dataset}
\begin{tabular}{lcc}
\hline

\makebox[0.08\textwidth][l]{Data Split} & \makebox[0.08\textwidth][c]{Questions} & \makebox[0.08\textwidth][c]{Answers*}\\
\hline
Training set & 8,782 & 5.13 \\
Test set & 102 & / \\
VSet1 & 4,963 & 5.06 \\
VSet2 & 52 & 10.40 \\
\hline
\multicolumn{3}{c}{\makecell[l]{
\specialrule{0em}{0.5pt}{0.5pt}
*The average answer number per question}}
\end{tabular}
\end{center}
\end{table}

We train our KEPR-based GenCQA model over the sole training set, and develop it on VSet1. In the self-contained experiments such as that in the ablation study, we follow the common practice~\cite{ma2021exploring, maskedLuo22} to test the model in VSet2. When comparing to the previous work in the main test, we report the performance released on the leaderboard\footnote{https://leaderboard.allenai.org/protoqa/submissions} of ProtoQA. It is obtained on the test set, where the questions are accessible, but the answers are undisclosed. Note that the evaluation process is implemented by the ProtoQA-evaluator, grounded on the submitted answers towards the disclosed questions.

\subsection{Evaluation Metrics}
We apply the official toolkit ProtoQA-evaluator to assess the GenCQA models. There are two kinds of metrics used for evaluation, including Ans@$k$ and Inc@$k$~\cite{boratko2020protoqa}. Both calculate the weighted accuracy ($\mathring{A}$cc.) given a certain truncation method. $\mathring{A}cc$ belongs to a task-specific evaluation scheme. It rewards a GenCQA model with higher scores if the popular-in-voting answers can be generated and highly ranked. We detail the calculation method of $\mathring{A}$cc in Appendix B.

The calculation of $\mathring{A}$cc in Ans@$k$ and Inc@$k$ is different from each other due to inconsistent truncation methods. For Ans@$k$, the answer list is truncated at the $k$-th answer, and the lower-ranked answers won't be considered for evaluation. For Inc@$k$, it is truncated at the $k$-th incorrect answer. We follow the canonical evaluation scheme of ProtoQA to set $k$ to 1, 3, 5 and 10 respectively in the separate experiments. Within all the metrics, Inc@3 is appointed as the most critical one, which accords with the tolerance of general users for errors. It is used as the gold standard for rating different submitted models on ProtoQA leaderboard.

\subsection{Implementation Details}
To speed up knowledge acquisition, we organize English Wiktionary dump into a hash map. The number of keywords used for retrieval is set to 2. During training, we optimize all our models with AdamW, where $\epsilon$ is set to 1e-6. Both generator and ranker are trained for one epoch with a batch size of 8. We respectively use GPT2, BART and T5 to construct different generators. All of them share the same hyperparameter settings, where the learning rate is set to 1e-5, and the warmup proportion is set to 0.025. When training the binary classifier for plausibility approximation, we set the learning rate to 3e-6, while the warmup proportion is set to 0.01. 

\subsection{Compared Models}

We utilize the fine-tuned GPT2 and BART as the baselines. The performance of GPT2 is reported in the pilot study of GenCQA on ProtoQA~\cite{boratko2020protoqa}. The performance of BART is reported on the ProtoQA leaderboard, which is accompanied with the easy-to-follow instructions of reproducing and fine-tuning prototypical BART~\cite{BART2021baseline}. In addition, we reproduce and enhance both GPT2 and BART to form their robust versions (GPT2-Robust and BART-Robust), where data cleaning is conducted on the training data (e.g., segmentation for compound answers and noise filtering), and hyperparameters are reset for model selection. Besides, the answer deduplication is used to diversify the output of GPT2 and BART.

We compare with Team Cosmic's work~\cite{T52021baseline} (T5-11B + Ranker), where the strong T5-11B possessing 11 billions of trainable parameters is used, and it is coupled with a ranker. We construct a T5-3B baseline (T5-3B-Robust). It is a robust version due to the use of data cleaning, fine-tuning and answer deduplication. 
In addition, SUDA NLP's work~\cite{GPT3Fewshot} (GPT3-175B Few-shot) is considered for comparison, which indicates the reasoning ability of large-scale language model.

Besides, we compare our models to a series of strong arts that use external knowledge, including:

\begin{itemize}
        \item BART+DPR~\cite{BARTDPR} which use DPR to build retriever. It acquires knowledge of relevant contexts from an external data source. The generator is built using BART.
        \item GPT2+ConceptNet~\cite{GPT2Concept} uses GPT2 to generate answers. It retrieves commonsense facts from ConceptNet~\cite{speer2017conceptnet}, and uses them to augment the input representations. DPR is also leveraged for retrieving relevant concepts in this art.
        \item T5-large+WordNet+Ranker~\cite{maskedLuo22} uses a large T5 network. It retrieves definition descriptions of keywords from WordNet~\cite{miller1995wordnet}, and incorporates them into the encoding process. Its ranker is constructed using DeBERTa~\cite{he2020deberta}.
\end{itemize}

\subsection{Main Results}

\begin{table*}[htbp]
\begin{center}
\caption{Performance comparison on the official (online) test set of ProtoQA.}
\label{tab3:mainresult}
\renewcommand{\arraystretch}{1.0}
\begin{tabular}{l|c|cccccc}
\hline
Method & \makebox[0.05\textwidth][c]{\textbf{Inc@3}} & \makebox[0.05\textwidth][c]{Inc@1} & \makebox[0.05\textwidth][c]{Inc@5} & \makebox[0.05\textwidth][c]{Ans@1} & \makebox[0.05\textwidth][c]{Ans@3} & \makebox[0.05\textwidth][c]{Ans@5} & \makebox[0.05\textwidth][c]{Ans@10}\\
\hline
GPT2 (baseline)~\cite{radford2019language} & 41.68 & 26.09 & 48.16 & 36.35 & 44.42 & 46.63 & 53.52\\
GPT2 + ConceptNet~\cite{GPT2Concept} & 41.23 & 26.32 & 48.60 & 36.76 & 41.06 & 45.84 & 55.28 \\
GPT3-175B Few-shot~\cite{GPT3Fewshot} & 47.63 & 35.76 & 53.53 & 47.08 & 50.11 & 51.93 & 56.94\\
GPT2 - Robust (reproduced) & 50.50 & 31.86 & 55.30 & 49.89 & 49.75 & 51.71 & 60.14\\
GPT2 - KEPR (ours) & 51.93 & 33.16 & 60.33 & 41.61 & 49.84 & 54.83 & 62.61\\
\hline
BART (baseline)~\cite{lewis2020bart} & 43.96 & 25.62 & 50.19 & 34.45 & 44.63 & 48.29 & 54.89\\
BART + DPR~\cite{BARTDPR}& 41.69 & 26.64 & 46.96 & 32.68 & 43.61 & 45.90 & 51.54 \\
BART - Robust (reproduced) & 51.69 & 35.46 & 57.28 & 45.52 & 49.95 & 53.03 & 61.60\\
BART - KEPR (ours) & 55.38 & 37.83 & 62.14 & 47.24 & 56.28 & 57.42 & 64.96\\
\hline
T5-11B + Ranker (baseline)~\cite{T52021baseline} & 54.15 & 43.59 & 55.59 & 56.00 & 60.98 & 60.66 & 56.15\\
T5-large + WordNet + Ranker~\cite{maskedLuo22} & 55.77 & 40.81 & 60.30 & 54.20 & 57.00 & 58.59 & 64.97\\
T5-3B - Robust (reproduced) & 57.05 & 38.55 & 64.12 & 55.87 & 58.28 & \textbf{62.65} & 66.44\\
\textbf{T5-3B - KEPR (ours)} & \textbf{60.91} & \textbf{45.72} & \textbf{67.87} & \textbf{60.52} & \textbf{61.14} & 62.13 & \textbf{69.42} \\
\hline
\multicolumn{8}{l}{\makecell[l]{
\specialrule{0em}{0.5pt}{0.5pt}
The weighted accuracy $\mathring{A}cc$ (\%) is used for evaluation, conditioned on two truncation schemes, i.e., Inc@$k$ and\\
Ans$k$. We follow the common practice to use Inc@$3$ is as the gold standard for rating different GenCQA models.
}}

\end{tabular}
\end{center}
\end{table*}

The performance of all the GenCQA models is shown in Table~\ref{tab3:mainresult}. It can be observed that KEPR yield substantial improvements, compared to the baselines and their robust versions. In addition, all the KEPR-based models outperform the state-of-the-art (SoTA) models which utilize the same PLMs. The performance gain is no less than 4.4\%, 11.5\% and 5.4\% at Inc@3 (gold standard) when GPT2, BART and T5-based SoTA are respectively taken for comparison.

More importantly, our BART-KEPR achieves the comparable performance at Inc@3, compared to T5-11B and T5-large based GenCQA models. Though, BART-KEPR merely contains 741M trainable parameters in total. By contrast, the T5-large generator contains 770M parameters, and its accompanying ELECTRA-based ranker additionally possesses 390M parameters. Besides, T5-11B obviously uses a huge number of parameters. Therefore, we suggest that our BART-KERP is vest-pocket and relatively practicable. It can be adopted by a potential user who intends to set up a baseline but lacks computational power.

\subsection{Ablation Study}

We carry out ablation experiments grounded on GPT2, BART and T5-3B over the secondary validation set VSet2 (i.e., the offline test set). Reverse ablation is conducted, where Knowledge Enhancement (KE) and answer Plausibility Ranking (PR) are coupled with the above PLM generators, respectively and independently. Further, they are jointly used (i.e., KEPR). Fig.~\ref{fig2:ablation} shows the experimental results, where the performance curves are plotted over the successive truncation points ($k\in$[1,10]) for both Ans@$k$ and Inc@$k$. The experimental results show that, in summary, the independent use of either KE or PR obtains minor improvements. By contrast, the joint use of them (i.e., KEPR) yields much more substantial improvements at all the metrics for every PLM baseline.

Another interesting finding is that the Ans@$k$ curve of KEPR in Fig.~\ref{fig2:ablation} rises steeply no matter whether $k$ equals to the minimum value ($k$=1) or largest value ($k$=10). This demonstrates that, on one hand, KEPR helps to push highly-weighted (i.e., popular in voting) answers up to the top of an answer list. On the other hand, it contributes to rolling a larger number of highly-weighted answers into a longer answer list. Though, this advantage cannot be absolutely attributed to KE, despite the performance curves of KEPR and KE seem to fluctuate synchronously in most cases. In fact, both KE and PR probably cause the effect, and each of them is indispensable. This can be demonstrated by the following findings:

\begin{figure}[th]
    \centering
	\includegraphics[width = \linewidth]{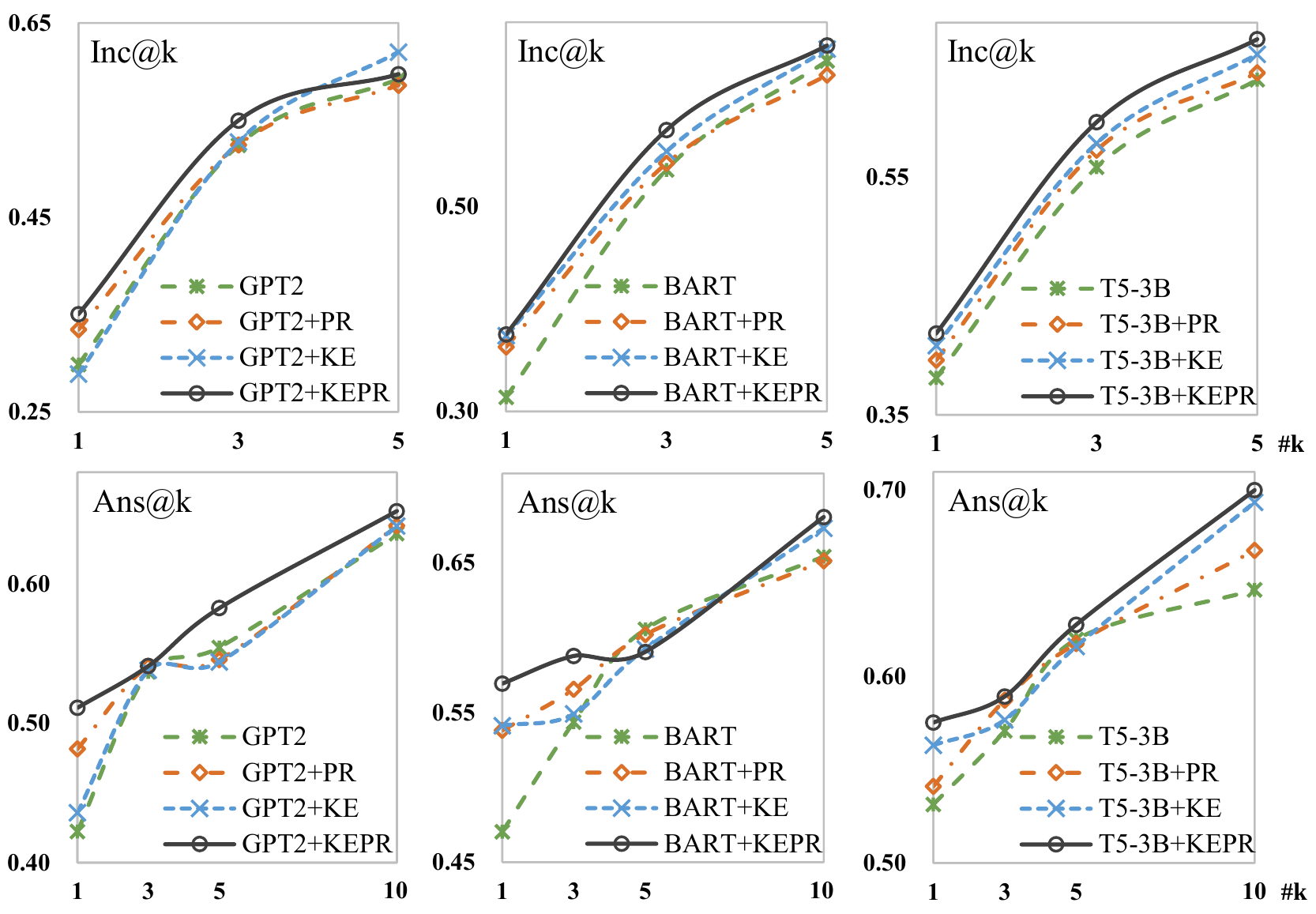}
	\caption{Ablation study for GPT2, BART and T5-3B on VSet2. The detailed performance metrics of each model are provided in Appendix C.}
	\label{fig2:ablation}
\end{figure}

\begin{itemize}
    \item When BART is considered as the baseline, PR causes performance degradation at Inc@$5$ ($k$=5). On the contrary, KE yields a relatively substantial improvement. It ensures the effectiveness of KEPR at Inc@$5$.
    \item When GPT2 is regarded as the baseline, KE results in performance reduction at Inc@$1$ ($k$=1). By contrast, PR obtains a significant improvement. Similarly, it allows KEPR to have a positive effect at Inc@$1$.
\end{itemize}

It can be additionally observed that both KE and PR cause performance reduction at Ans@$5$ ($k$=5) no matter which PLM is used as the baseline. It is because that PLM baselines overfit the mostly short answer lists in the training and validation sets. Specifically, the length of ground-truth answer lists in the training set is 5.13 on average, while 5.06 in the validation set VSet1. By contrast, the average length of answer lists in VSet2 is about 10.40. The PLMs are trained to attain the most perfect states over the shorter answer lists, though uncertainty is increased when a longer answer list needs to be generated during test. By contrast, when KE and PR are used, the whole models (PLMs+KE and PLMs+PR) can be trained to converge to the globally optimal solutions on the longer answer lists, instead of locally optimal solutions. They obviously adapt to the test data better, although this is at the expense of performance reduction for the top-$5$ answers.

It is noteworthy that, during the training of PLM baselines, arbitrarily enlarging the maximum length of the generated answer lists is ineffective. It is because that the length of ground-truth answer lists is fixed, and thus the estimated cross-entropy loss (between the generated answers and ground truth) for backpropagation will not be changed significantly, along with the modification of maximum length. By contrast, the effects of KE and PR are active when the maximum length is set to a larger number. The answer deduplication module in KE dynamically influences the loss of top-$1$, $3$ or $5$ answers (by redundancy elimination and novel answer replacement). Besides, PR serves as the post-treatment which separately refines the rankings of not only top-$5$ cases but top-$10$.

\begin{figure}[t]
    \centering
	\includegraphics[width =0.9\linewidth]{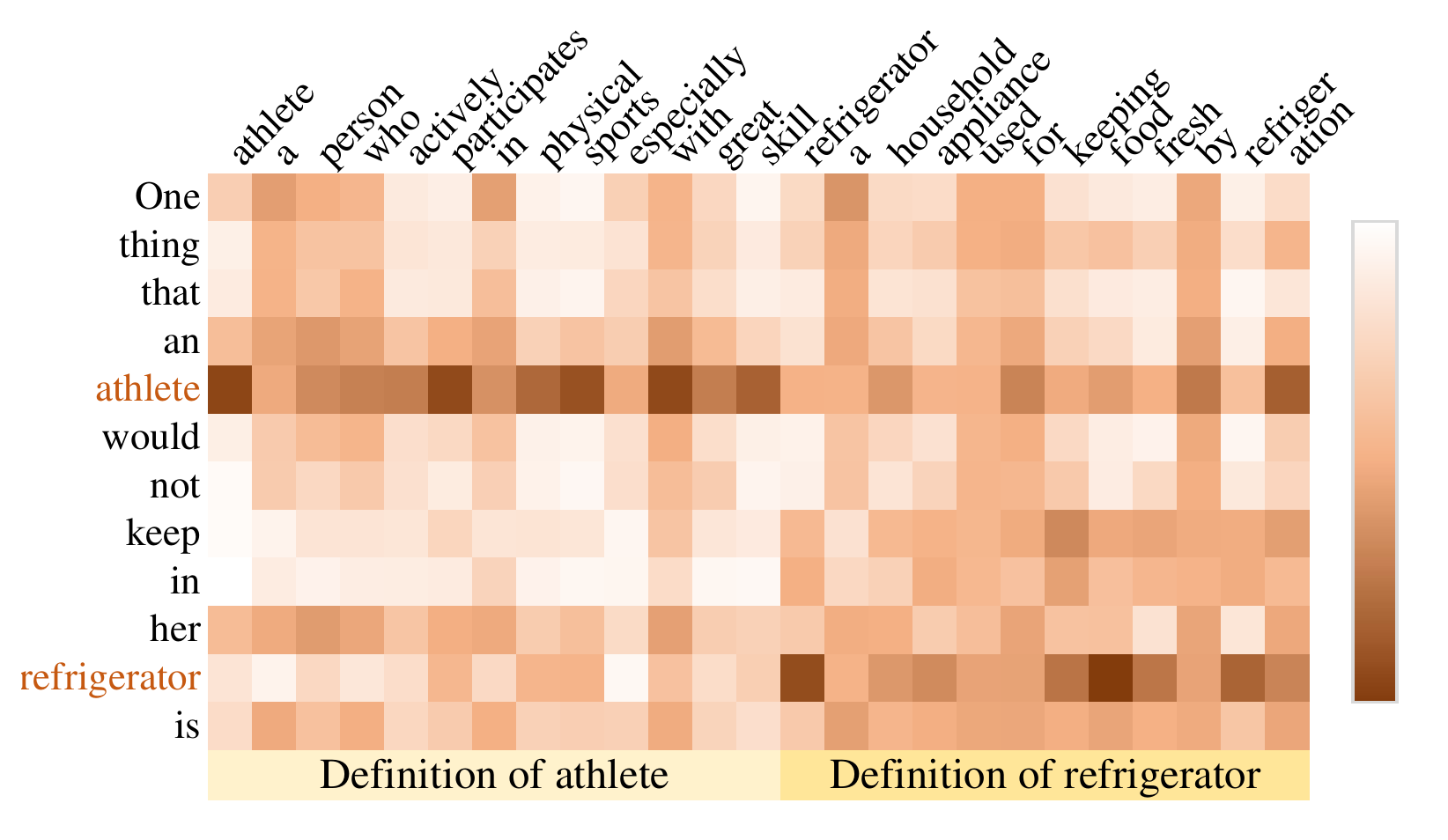}
	\caption{Attention distribution diagram (BART+KE).}
	\label{fig3:attention}
\end{figure}

\begin{table}[t]
\begin{center}
\caption{Answer generation examples.}
\label{tab4:case study}
\renewcommand{\arraystretch}{1.0}
\begin{tabular}{ll}
\hline
\textbf{Question \#1} & \makecell[l]{ One thing that an \textit{\underline{\textbf{athlete}}} would not keep in\\
her \textit{\underline{\textbf{refrigerator}}} is}\\
\hline
Knowledge & \makecell[l]{
\textbf{\textit{athlete}}: a person who actively participates in \\ physical sports, especially with great skill.\\
\textbf{\textit{refrigerator}}: a household appliance used for\\  keeping food fresh by refrigeration.\\
}\\
\hline
BART & \makecell[l]{
beer, ice cream, alcohol
}\\
BART+KE & \makecell[l]{
beer, ice cream, chocolate, clothes, socks
}\\
\hline
\specialrule{0em}{4pt}{4pt}
\hline
\textbf{Question \#2} & One thing a \textit{\underline{\textbf{monk}}} \textit{\underline{\textbf{probably}}} would not own is\\
\hline
Knowledge & \makecell[l]{
\textbf{\textit{monk}}: A male member of a monastic order \\ who has devoted his life for religious service.\\
\textbf{\textit{probably}}: In all likelihood. \\
}\\
\hline
BART & \makecell[l]{
car, cell phone, money}\\
BART+KE & \makecell[l]{car, cell phone, money, gun, jewelry, wife\\
}\\
\hline
\\
\multicolumn{2}{l}{\makecell[l]{
The answers generated by BART and its enhanced version\\
(+KE) are considered.
}}
\end{tabular}
\end{center}
\end{table}

\subsection{Effects of Knowledge Enhancement}
KE contributes to the diversification of answers. Let's consider the BART-based GenCQA model which is merely coupled with KE. It additionally recalls 34\% different classes of answers (see the examples in Table~\ref{tab4:case study}), in terms of our verification on VSet2. This contributes to the improvement of Ans@$12$ with a rate of up to about 14.2\%. The positive effects of KE benefit from the increased intention weights of keywords, i.e., the ones obtained by interaction between questions and definition descriptions of keywords. Fig.~\ref{fig3:attention} shows an example of attention distribution diagram. In Appendix D, we present the method for creating attention heat maps.

\subsection{Reliance of Knowledge Enhancement}
KE heavily relies on the qualified knowledge items. All the pretreatments (TF-IDF keyword extraction, DPR and the setting of number) may influence the quality.   

\subsubsection{Reliability of Keyword Extraction}
We extract top-$m$ keywords for each question conditioned on the statistical distribution feature TF-IDF. The words possessing higher TF-IDF scores will be adopted as the keywords. We verify the reliability of the keywords in a separate experiment. Specifically, we randomly select 50 questions from the validation set VSet1, and use the aforementioned statistical approach to extract keywords. Meanwhile, we manually annotate the keywords, and rank them according to their importance. There are 2.3 keywords on average annotated for each question, and the maximum number is 4. We determine the extracted keywords to be correct if the agreement is reached conditioned on the annotation results. On this basis, we verify the Macro-average accuracy for the top-$m$ ($m\leq$4) TF-IDF extracted keywords. The performance is 0.88, 0.80, 0.69 and 0.56 for top-$1$, $2$, $3$ and $4$ keywords, respectively.

\begin{figure}[t]
    \centering
	\includegraphics[width = 0.8\linewidth]{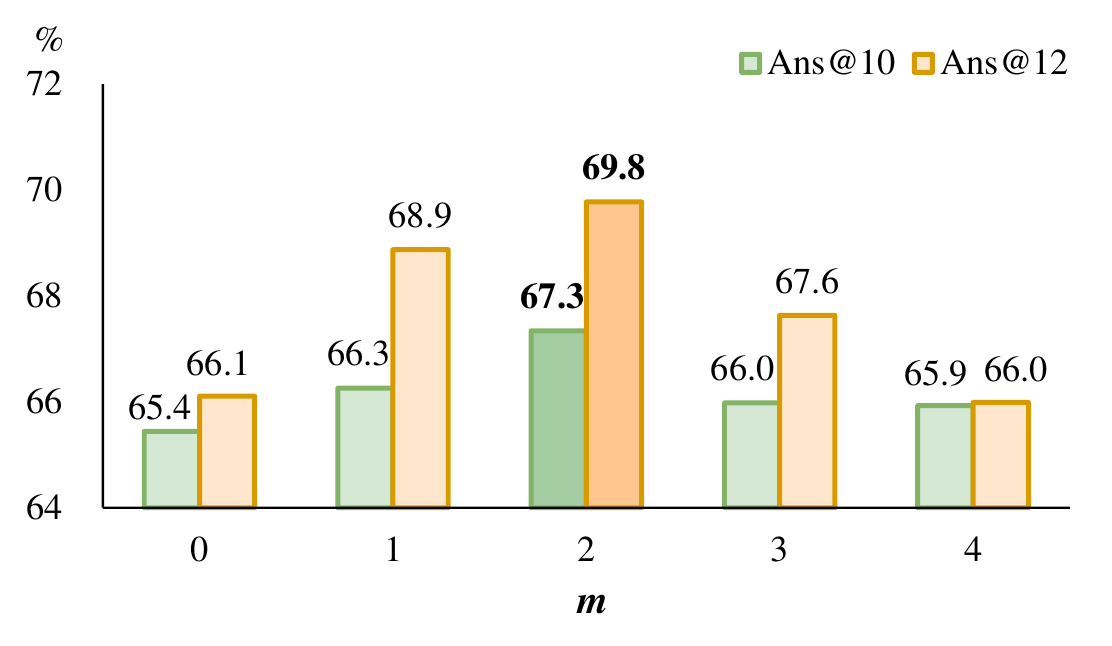}
	\caption{Effects of $m$ on Knowledge Enhanced Answer Generation. The performance of best $m$ is in \textbf{bold}.}
	\label{fig:conceptnum}
\end{figure}

\subsubsection{Reliability of DPR}
We use DPR to determine the semantic similarity between keyword-based queries and definition descriptions for selecting relevant descriptions (Section III.A). We evaluated DPR's reliability for knowledge acquisition by manually verifying 50 keywords and found an accuracy of 0.86, significantly higher than simply using the primary definition~\cite{xu2021fusing} (accuracy of 0.66). Note that the primary definition refers to the top-1 placed definition in Wiktionary. 

\subsubsection{Number of Knowledge Items}

KE is used to increase the diversity of answers by covering a wider range of answer categories. To determine the optimal setting for the number of available knowledge items $m$, we conduct a preliminary evaluation of enhanced baselines (PLMs+KE) at different values of $m$. Ans@10 and Ans@12 serve as evaluation metrics as they reflect overall recall. The performance of using different numbers of knowledge items was evaluated and shown in Fig.~\ref{fig:conceptnum}. Results indicate that using too few ($m=0$ or $m=1$) or too many knowledge items for KE decreases the accuracy. Using too few knowledge items is ineffective in capturing various answer classes, while using too many introduces noise from incorrect keywords. Therefore, we set $m$ to 2 to balance accuracy and diversity of knowledge answers.

\subsection{Constructing a Proper PR Corpus}

To rank the generated answers, we use a plausibility scoring model based on the likelihood of an answer being completely correct and reasonable. We train an ELECTRA-based binary classifier using a corpus of positive and negative answers. For selecting positive instances, we choose the top-$n$ answers from the ground-truth list in descending order of popularity. We experiment with different values of $n$ and find that a value of 2 yields the best performance. The detailed experiment result and case study are provided in Appendix E.

\section{Conclusion}
We propose a KEPR approach to strengthen the PLMs-based GenCQA models. Experimental results demonstrate that our approach yields substantial improvements, and the resultant GenCQA models outperform the state of the art. In addition, KEPR comprises three parts, including knowledge retriever, answer generator and plausibility ranker. Each of them is designed to be a general component for GenCQA tasks. However, there is only one GenCQA corpus publicly released for evaluation. Therefore we fail to verify and guarantee the generality on multiple datasets. If any other GenCQA dataset is publicly available, we will report the generality on them, so as to provide a reference for future research.

\bibliographystyle{IEEEtran}
\bibliography{cite}

\end{document}


\title{\huge{KEPR: Knowledge Enhancement and Plausibility Ranking for Generative Commonsense Question Answering}}

\author{
\IEEEauthorblockN{Zhifeng Li}
\IEEEauthorblockA{
\textit{School of Computer Science and Technology} \\
\textit{Soochow University}, Suzhou, China \\
li\_2hifeng@outlook.com}
\\
\IEEEauthorblockN{Yifan Fan}
\IEEEauthorblockA{
\textit{School of Computer Science and Technology} \\
\textit{Soochow University}, Suzhou, China \\
yifanfannlp@gmail.com}
\and 
\IEEEauthorblockN{Bowei Zou}
\IEEEauthorblockA{
\textit{Institute for Infocomm Research} \\
\textit{A*STAR}, Singapore \\
zou\_bowei@i2r.a-star.edu.sg}
\\
\IEEEauthorblockN{Yu Hong\textsuperscript{\Letter}}
\IEEEauthorblockA{
\textit{School of Computer Science and Technology} \\
\textit{Soochow University}, Suzhou, China \\
tianxianer@gmail.com
}}

\maketitle

\section*{Appendix}

\subsection{Question Rewriting}
All the ProtoQA questions begin with a specific prefix. Some of the prefixes appear as the commonly-used question words like ``{\em what}'', which are also known as the heads of interrogative sentences (question heads for short). Others are constituted by a series of chatty prologues like ``{\em tell me something}''. Table~\ref{tab5: mapping table} shows the frequently-occurred question heads. They account for about 98 percent of the ProtoQA training set.

Miscellaneous heads are misleading for perceiving question intentions. For example, the token ``{\em name}'' in the question head of ``{\em name an}'' may correspond to the intention of pursuing an name entity, although the head actually raises the question of naming some objects, facts or events. To avoid misunderstanding, we simplify the question heads, and rewrite the questions using the unified patterns, which are formatted as below: 
\begin{equation}
\small
\mathrm{head <ctt> is}
\end{equation}
where, $head$ denotes the new question head, which is assigned manually in terms of the original question head. There are three options in total for assigning a $head$, including ``{\em one}'', ``{\em one thing}'' or ``{\em one way to tell}''. The character $ctt$ refers to the question content, which is extracted from the original question by simply pruning the original head. Accordingly, we set up a small mapping table between original question heads and rewriting patterns, as shown in Table~\ref{tab5: mapping table}. 

In practice, we rewrite all the ProtoQA questions by looking up patterns in the mapping table, and substituting the real question contents into $<ctt>$ of the corresponding patterns.

\subsection{Weighted Accuracy}
The weighted accuracy ($\mathring{A}$cc.) belongs to a task-specific evaluation scheme, which is different from the commonly-used traditional accuracy. Assume the length of the answer list before the truncation point is $\mathring{l}en$, the accuracy is calculated as follows: 
\begin{equation}
\small
\begin{array}{l}
\begin{aligned}
&\boldsymbol{\mathring{A}cc.}=\frac{\sum_{i=1}^{\mathring{l}en}\sum_{j=1}^{N_c}\boldsymbol{r}_{ij}}{\sum_{i=1}^{\mathring{l}en}\boldsymbol{\mathring{R}}_i}
\end{aligned}
\end{array}
\label{formula:process}
\end{equation}
where, $r_{ij}$ denotes the weight rewarded to the $i$-th answer $a_i$ (i.e., the generated answer that is ranked in the $i$-th place). If $a_i$ belongs to the $j$-th ground-truth answer class $C_j$, the crowd-sourced weight of $C_j$ will be rewarded to $a_i$, otherwise 0 is rewarded. $\mathring{R}_i$ is the ideal reward for $a_i$. It is equivalent to the weight of the $i$-th popular answer class. $N_c$ is the number of answer classes. 

\begin{table}[t]
\begin{center}
\caption{The mapping table of question rewriting.}
\begin{tabular}{lcl}
\hline
Original Head (Prefix) & Proportion & Patterns\\
\hline
\textit{name something} & 44.53\% & \textit{one thing} \textless\textit{ctt}\textgreater \textit{is}\\
\textit{name a} & 29.08\% & \textit{one} \textless\textit{ctt}\textgreater \textit{is}\\
\textit{what} & 6.08\% & \textit{one thing} \textless\textit{ctt}\textgreater \textit{is}\\
\textit{name an} & 5.78\% & \textit{one} \textless\textit{ctt}\textgreater \textit{is}\\
\textit{name} & 5.16\% & \textit{one} \textless\textit{ctt}\textgreater \textit{is}\\
\textit{tell me something} & 2.41\% & \textit{one thing} \textless\textit{ctt}\textgreater \textit{is}\\
\textit{which} & 1.59\% & \textit{one} \textless\textit{ctt}\textgreater \textit{is}\\
\textit{tell me a} & 1.25\% & \textit{one} \textless\textit{ctt}\textgreater \textit{is}\\
\textit{tell me} & 1.03\% & \textit{one} \textless\textit{ctt}\textgreater \textit{is}\\
\textit{give me a} & 0.35\% & \textit{one} \textless\textit{ctt}\textgreater \textit{is}\\
\textit{tell me an} & 0.33\% & \textit{one} \textless\textit{ctt}\textgreater \textit{is}\\
\textit{how can you tell} & 0.18\% & \textit{one way to tell} \textless\textit{ctt}\textgreater \textit{is}\\
others & 2.22\% & \textit{Q:} \textless\textit{ctt}\textgreater \quad \textit{A:}\\
\hline
\multicolumn{3}{l}{\makecell[l]{
\specialrule{0em}{0.5pt}{0.5pt}
``others'' refers to the infrequently-occurred question heads.
}}
\end{tabular}
\end{center}
\label{tab5: mapping table}
\end{table}

\subsection{Detail Metrics for Ablation Study}
Table~\ref{tab2:ablation_full} lists the detailed metrics of GPT2, BART and T5 for Ablation Study. Where KE indicates knowledge enhancement for answer generation and PR denotes plausibility ranking for generated answer. 

\subsection{Interactive Attention Between Question and Knowledge}
As mentioned in Section IV.G, experimental results show that KE helps to improve the versatility, i.e., recalling a larger number of answer classes compared to the baseline BART. In addition, we suggest that this advantage most probably benefits from the highlighted attention weights of keywords in questions, instead of the common or meaningless words (e.g., stop words).
In this appendix, we present the method of creating attention heat maps between questions and knowledge, so as to support the insightful study of inner attention distributions. Similarly, we take BART+KE as the example in the presentation.

Specifically, we feed a question and corresponding knowledge items (i.e., the definition descriptions retrieved using keywords) into BART+KE (i.e., a medium-level case in our model family). When BART encoding is terminated, we conduct element-wise attention accumulation over all attention heads among all the transformer layers. On this basis, we stripe the self-attentions and organize the knowledge-to-question attentions in the order of their token sequences.

\begin{table*}[t]
\begin{center}

\caption{Detailed Performance of Models for Ablation study.}
\label{tab2:ablation_full}
\begin{tabular}{l|c|ccccccc}
\hline

Method & \makebox[0.05\textwidth][c]{\textbf{Inc@3}} & \makebox[0.05\textwidth][c]{Inc@1} & \makebox[0.05\textwidth][c]{Inc@5} & \makebox[0.05\textwidth][c]{Ans@1} & \makebox[0.05\textwidth][c]{Ans@3} & \makebox[0.05\textwidth][c]{Ans@5} & \makebox[0.05\textwidth][c]{Ans@10} & \makebox[0.05\textwidth][c]{Ans@12}\\

\hline
GPT2 & 52.43 & 29.87 & 59.24& 42.23 & 53.68 & 55.44 & 63.59 & 64.92\\
+ PR & 52.48 & 33.46 & 58.61 & 48.16 & 54.08 & 54.55 & 64.17 & 64.92\\
+ KE & 52.78 & 28.89 & \textbf{62.03} & 43.57 & 53.80 & 54.40 & 64.14 & \textbf{66.67}\\
+ KEPR & \textbf{54.98} & \textbf{35.05} & 59.73 & \textbf{51.11} &\textbf{54.11} & \textbf{58.26}& \textbf{65.22} & \textbf{66.67} \\
\hline
BART & 53.60 & 31.39 & 64.23 & 47.04 &54.37 & \textbf{60.56} & 65.45 & 66.11\\
+ PR & 54.22 & 36.31 & 62.85 & 53.82 & 56.56 & 60.23 & 65.15 & 66.11\\
+ KE & 55.39 & 37.40 & 65.38 & 54.13 & 54.91 & 59.25 & 67.34 & \textbf{69.78}\\
+ KEPR & \textbf{57.48} & \textbf{37.54} & \textbf{65.73} & \textbf{56.94} & \textbf{58.79} & 59.06& \textbf{68.09} & \textbf{69.78}\\
\hline
T5-3B & 53.60 & 31.39 & 64.23 & 47.04 &54.37 & \textbf{60.56} & 65.45 & 66.11\\
+ PR & 54.22 & 36.31 & 62.85 & 53.82 & 56.56 & 60.23 & 65.15 & 66.11\\
+ KE & 55.39 & 37.40 & 65.38 & 54.13 & 54.91 & 59.25 & 67.34 & \textbf{69.78}\\
+ KEPR & \textbf{57.48} & \textbf{37.54} & \textbf{65.73} & \textbf{56.94} & \textbf{58.79} & 59.06& \textbf{68.09} & \textbf{69.78}\\
\hline
\end{tabular}

\end{center}

\end{table*}

\subsection{Constructing a Proper PR Corpus}

\begin{figure}[t]
    \centering
	\includegraphics[width = 0.8\linewidth]{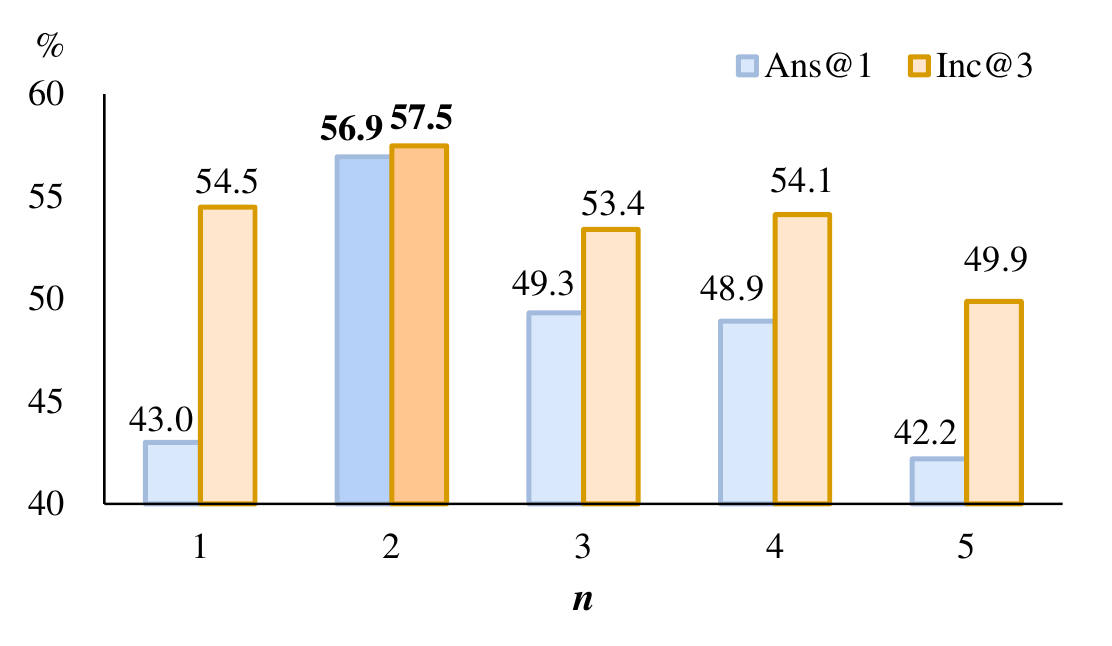}
	\caption{Performance of BART+PR trained on different PR corpora that were built using different settings of $n$.}
	\label{fig6:answernum}
\end{figure}

Our ranker sorts the generated answers grounded on plausibility scores. Plausibility is estimated in the process of determining how possible an answer is positive. The ELECTRA-based binary classifier is constructed and trained to implement plausibility estimation. In order to train such a classifier, we build a corpus that contains positive answers of different qualities, as well as randomly-selected negative answers. In this case, it is crucial to select the proper number of positive instances, from the ground-truth answer list in the top-down manner. Note that the ground-truth list is formed in descending order of the percentage of the vote (i.e., popularity).

Given a question and the accompanying ground-truth answer list, we select top-$n$ answers as the positive instances to construct the corpus. We verify the utility of a series of corpora obtained using different settings of $n$. The utility is evaluated by the performance of a PLM-based generator that is coupled with PR (PLM+PR). Similarly, we consider BART+PR in the corresponding experiments. Fig.~\ref{fig6:answernum} shows the performance obtained on the validation set VSet2.

It can be observed that the performance is worse when $n$ is set to 1. We believe that, in this case, the model of BART+PR suffers from data sparsity. The current corpus merely contains 7,962 positive examples and the same amount of negative instances. In addition, we find that the performance gradually degrades when $n$ is larger than 2. It is because that a larger number of unpopular positive answers are involved into the corpus. Such answers appear as the all-purpose but less relevant cases. For example, the unpopular answer class ``{\em clothing}'' in a) can be used as the dry answers in b).


\begin{itemize}
    \item [a)] {Question}: {\em Name \underline{something} that an \textbf{athlete} would not keep in her \textbf{refrigerator}}?\\
    Answer Class (``{\em clothing/shoe}''): {\em gloves}, {\em clothes}, {\em shoe}, ...
\end{itemize}

\begin{itemize}
    \item [b)] {Question}: {\em Name \underline{something} that \textbf{cats} like to \textbf{rub} up against}?
\end{itemize}

As a result, it is difficult to distinguish the unpopular positive answers from negative. When a larger amount of unpopular answers are involved into the training set, the classifier will fail to learn the appropriate plausibility computing mode. Therefore, we set $n$ to 2 in our experiments, with the aim to ensure the quantity and quality of positive instances during the construction of the PR corpus.